\documentclass[publish,conference,paper]{conferencearticle}
\usepackage[dvips]{graphicx}
\usepackage{epsfig}

\begin{document}

\pagestyle{conferencestyle}

\title{Analysis of the Production Strategy of Mask Types in the COVID-19 Environment}
\author{Xiangri LU, Zhanqing WANG, Hongbin MA}
\address{School of Automation, Beijing Institute of Technology, Haidian District, Beijing}
\markboth{Xiangri LU, Zhanqing WANG, Hongbin MA}{Analysis of the Production Strategy of Mask Types in the COVID-19 Environment}
\maketitle

\begin{abstract}
\noindent Abstract. Since the outbreak of the COVID-19 in December 2019, medical protective equipment such as disposable medical masks and KN95 masks have become essential resources for the public. Enterprises in all sectors of society have also transformed the production of medical masks. After the outbreak, how to choose the right time to produce medical protective masks, and what type of medical masks to produce will play a positive role in preventing and controlling the epidemic in a short time. In this regard, the evolutionary game competition analysis will be conducted through the relevant data of disposable medical masks and KN95 masks to determine the appropriate nodes for the production of corresponding mask types. After the research and analysis of the production strategy of mask types, it has a positive effect on how to guide the resumption of work and production.
\end{abstract}


\section{Introduction}
The COVID-19 is a highly infectious disease following SARS and Middle East Respiratory Syndrome. According to official reports, the effective way of prevention and control is vaccination. Before the vaccine is successfully developed, medical masks can effectively prevent the spread of the virus. Masks are an important line of defense to prevent respiratory infectious diseases and can reduce the risk of COVID-19[1] infection. Masks can not only prevent the patient from spraying droplets, reduce the amount and speed of droplets, but also block the virus-containing droplet nuclei and prevent the wearer from inhaling, which has a good effect on controlling the epidemic. According to statistics related to masks, the effective protection time of ordinary disposable medical masks is 4 hours, and the effective protection time of KN95 masks is 15 hours. The production of one KN95 mask[2,3] can produce 3 ordinary disposable medical masks, and it can be judged that the production of masks conforms to the evolutionary game competition model[4,5]. Starting from the production mask market system[6,7,8,9], the adjustment behavior of the mask production group is regarded as a dynamic system, which constitutes a macro model with a micro analysis basis, so it can more truly reflect the diversity and complexity of the actors, and can be a macro model. Provide a theoretical basis for regulating group behavior in the production of masks.
First of all, this article analyzes how to supply masks to front-line medical staff from factors such as inventory support capacity, production rate of masks, and use efficiency of different types of masks to meet the daily consumption of three masks for all regional personnel. Secondly, this paper designs the algorithm based on the evolutionary game algorithm and the principle of the competition mechanism[10,11] between organisms, and uses the Runge-Kutta method[12] to calculate the numerical solution of the mathematical model of mask production[13] under the competition mechanism.

\section{The Design of Production Model of Mask Type}
As there are too many factors affecting the production of masks, only some factors are considered here, and factors such as consumer purchasing power are temporarily not considered. Assume a regional city with a population of 3 million, with 30,000 health technicians. According to the regulations of relevant national departments, the amount of medical emergency reserve materials should not be less than 10 days. Assuming that all 10,000 front-line medical personnel wear KN95 masks, and the rest wear disposable protective masks, the daily consumption of three masks per person is one time. The stock of masks is 600,000, and the number of KN95 masks is 300,000.
The second is to define the efficiency of producing different mask types. According to market data, the time to produce a KN95 mask can produce 3 ordinary disposable medical masks, that is, $r_{1}=1,r_{2}=3$; ordinary disposable masks and KN95 masks The maximum production volume is three masks per person per day in the area, that is, $n_{1}=n_{2}=900$million pieces; theoretically, the effective protection time of ordinary disposable medical masks is 4 hours, and the effective protection time of KN95 masks is 15 hours. In the effective protection state, the number of KN95 masks consumed per unit time is $s_{1}=0.27$ times that of ordinary disposable medical masks. The unit time consumption of ordinary disposable medical masks is $s_{2}=3.75$times that of KN95 masks.After the mask type production background model is designed, a mathematical model is established using the population competition model.

\section{BASIC IDEA}
When the two medical resources, ordinary disposable medical masks and KN95 masks, compete for social resources and living space, the common outcome is the extinction of weak competitiveness, and the strong competitiveness reaches the maximum capacity allowed by the environment. Analyze the conditions that produce various endings. Equations (1) and (2) are the differential equations of KN95 masks and ordinary disposable medical masks in the evolution of the amount of medical resources.
\begin{equation}
\frac{d x}{d t}=r_{1} x\left(1+\frac{x}{n_{1}}+s_{1} \frac{y}{n_{2}}\right)
\end{equation}

\begin{equation}
\frac{d y}{d t}=r_{2} y\left(1+\frac{y}{n_{2}}+s_{2} \frac{x}{n_{1}}\right)
\end{equation}

Where, $x(t),y(t)$ are the number of KN95 masks and ordinary disposable medical masks that change over time, $r_{1},r_{2}$ are their production efficiency, and $n_{1},n_{2}$ are the maximum output. $s_{1}$ is the magnification of ordinary disposable medical masks consumed by KN95 masks per unit time, and $s_{2}$ is the magnification of ordinary disposable medical masks consumed per unit time of KN95 masks.
The fourth-order Runge-Kutta method is used to solve the differential equation of the number of KN95 masks and ordinary disposable medical masks over time. First, iteratively calculate the number of KN95 masks and ordinary disposable medical masks shown in equations 3 and 4, and then obtain the numerical solution iterative equation for the number of high-precision fourth-order Runge-Kutta method KN95 masks and ordinary disposable medical masks.such as Equation 5 and Equation 6 Shown.

\begin{equation}
x_{n+1}=x_n+h f(x_n, y_n)
\end{equation}
\begin{equation}
y_{n+1}=y_n+h f(x_n, y_n)
\end{equation}
Where $x_{n}$ represents the number of KN95 masks at time n, $y_{n}$ represents the number of ordinary disposable medical masks at time n, $h$ represents the step length of the iterative equation, and $f(x_{n}, y_{n})$ represents the rate of change in the number of KN95 masks and ordinary disposable medical masks .
\begin{equation}
\left\{\begin{array}{l}
x_{n+1}=x_{n}+\frac{h}{6}\left(k_{1}+2 k_{2}+2 k_{3}+k_{4}\right) \\
k_{1}=f\left(x_{n}, y_{n}\right) \\
k_{2}=f\left(x_{n}+\frac{h}{2}, y_{n}+\frac{h}{2} k_{1}\right) \\
k_{3}=f\left(x_{n}+\frac{h}{2}, y_{n}+\frac{h}{2} k_{2}\right) \\
k_{4}=f\left(x_{n}+h, y_{n}+h k_{3}\right)
\end{array}\right.
\end{equation}
\begin{equation}
\left\{\begin{array}{l}
y_{n+1}=y_{n}+\frac{h}{6}\left(k_{1}+2 k_{2}+2 k_{3}+k_{4}\right) \\
k_{1}=f\left(x_{n}, y_{n}\right) \\
k_{2}=f\left(x_{n}+\frac{h}{2}, y_{n}+\frac{h}{2} k_{1}\right) \\
k_{3}=f\left(x_{n}+\frac{h}{2}, y_{n}+\frac{h}{2} k_{2}\right) \\
k_{4}=f\left(x_{n}+h, y_{n}+h k_{3}\right)
\end{array}\right.
\end{equation}
Where $k_{1}$, $k_{2}$, $k_{3}$,$k_{4}$ are undetermined parameters.
The fourth-order Runge-Kutta method is a high-precision single-step method; the numerical stability is good; you only need to know the first-order derivative, and you do not need to explicitly define or calculate other higher-order derivatives; just give $x_{n},y_{n}$ to calculate $x_{n+1},y_{n+1}$; This article uses MATLAB to program the Runge Kuta method. The MATLAB environment already has an integrated script for the Runge Kuta method, which is easy to implement in programming.
Through the calculation of the numerical solution of the fourth-order Runge-Kutta method, the output changes of KN95 masks and ordinary disposable medical masks in a part of time with a step length of 0.1 are obtained, as shown in Table 1.
\begin{table}[thb]
  \begin{center}
    \caption{Evolution of the Number of Two Types of Masks}
    \begin{tabular}[t]{cccccc}
    \hline
    Evolution Time & Number of KN95 & Number of disposable masks\\
	\hline
    0.0 & 30.000 & 60.000\\
    0.1 & 32.972 & 76.128\\
    0.2 & 36.207 & 95.648\\
	0.3 & 39.719 & 118.820\\
    0.4 & 43.522 & 145.711\\
    0.5 & 47.626 & 176.117\\
	0.6 & 52.043 & 209.493\\
    0.7 & 56.781 & 244.940\\
    0.8 & 61.851 & 281.242\\
	0.9 & 67.261 & 316.971\\
	1.0 & 73.023 & 350.663\\
    \hline
    \end{tabular}
    \label{table}
  \end{center}
\end{table}
\section{SIMULATION STUDIES}
In this part, we will give the production background model simulation of mask types, and analyze the best production time nodes of KN95 masks and ordinary disposable medical masks and how to adjust the two types of masks under the production background model of each mask type. Output can play a positive role in epidemic prevention.
\subsection{situation 1}
If the development of the regional mask industry is at an intermediate level, the production background of the mask type is simulated in accordance with the description in Section 2. The simulation results are shown in Figure 1, Figure 2.
Figure 1 shows the evolutionary game confrontation between KN95 masks and ordinary disposable medical masks. 1.8 time units before the outbreak, the output of disposable masks had an advantage in the market. At this time, the output of KN95 masks was steadily increasing. Between the second time unit and the 2.8th time unit, the number of KN95 masks was still rising, but the market share of disposable masks began to decline. After the 2.8th time unit, the number of KN95 masks has an overwhelming advantage, and the market share of disposable masks will be gradually replaced by KN95 masks.
From the above analysis, it can be seen that if KN95 mask manufacturers want to capture market share at the least cost, they must start to increase the production of KN95 masks after the second time unit. After the fifth time unit, disposable masks have been unable to support regional fights against the epidemic. The protective capabilities and timeliness of KN95 masks have played an important role. At this time, KN95 masks in the region are close to saturation. Manufacturers should adjust their production strategies according to market demand. Slow down the production speed, and produce the corresponding number of masks, so as to avoid excessive masks and waste.
\begin{figure}[t]
\centering
\includegraphics[width=6cm]{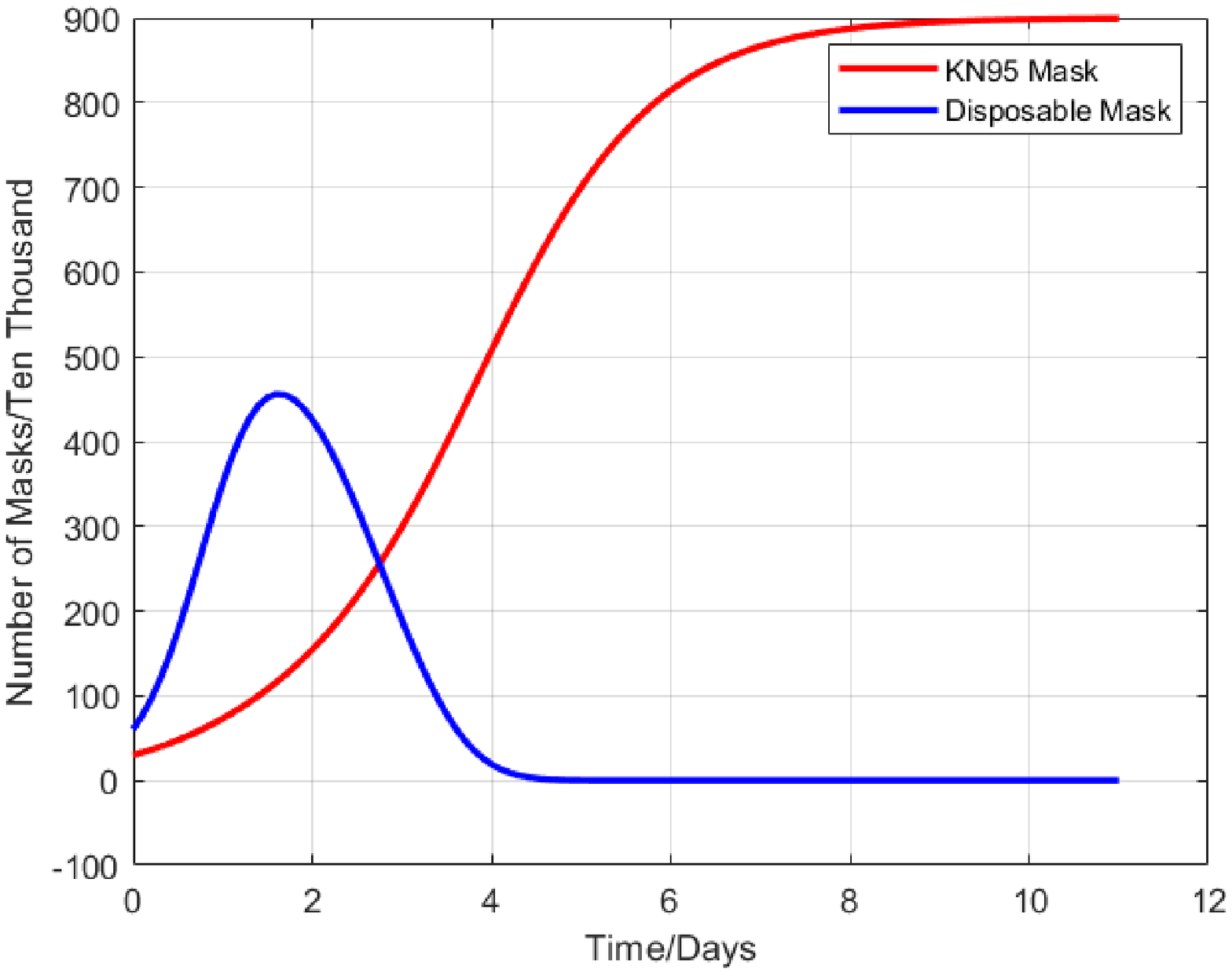}
\caption{Evolutionary Game of Two Types of Masks}
\label{fig}
\end{figure}

\begin{figure}[t]
\centering
\includegraphics[width=6cm]{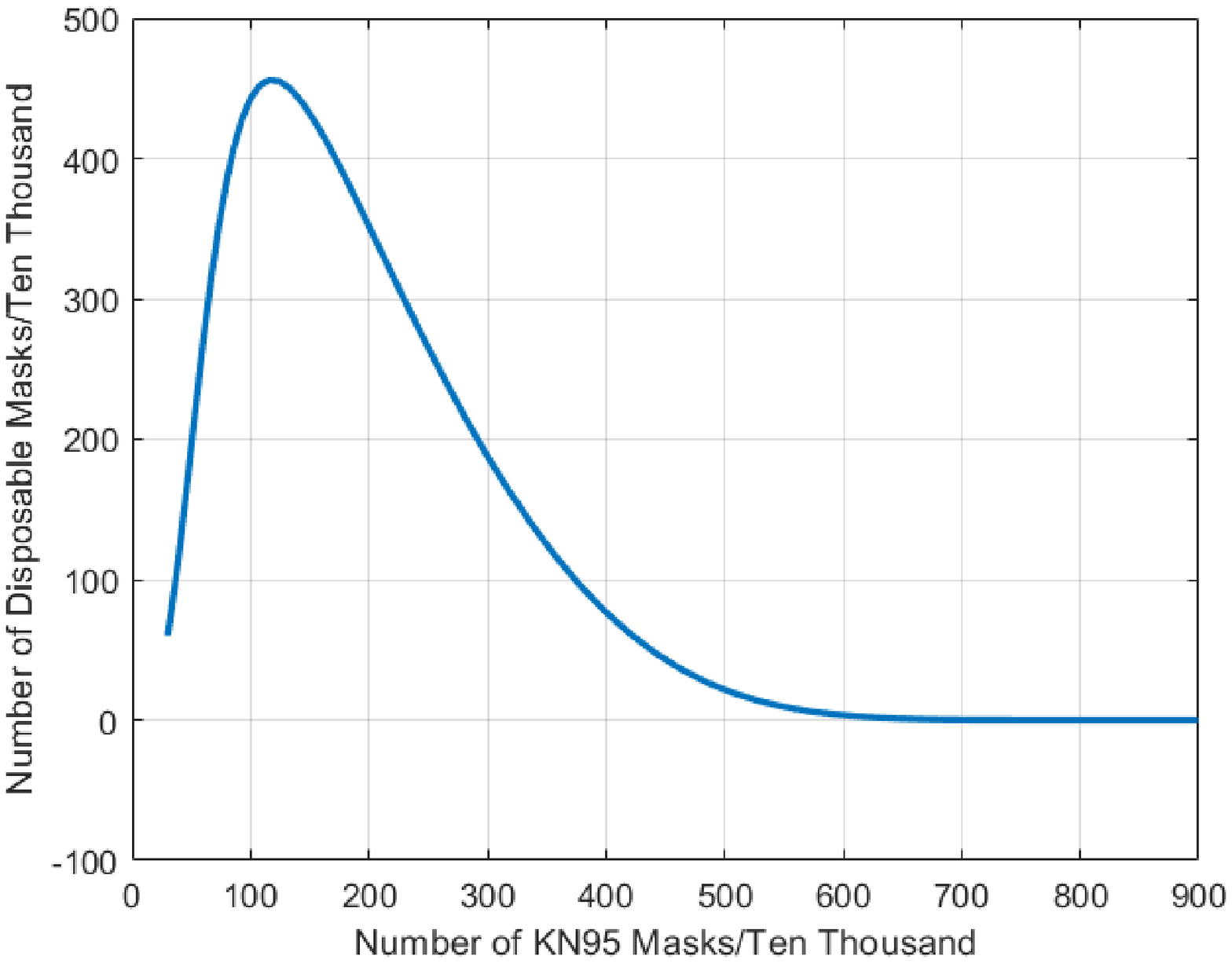}
\caption{Two Types of Masks Number Comparison}
\label{fig}
\end{figure}

\subsection{situation 2}
If the development of the regional mask industry is at a high level. The ability of regional masks to produce KN95 masks is higher than the medium level, that is, $r_{1}=2,r_{2}=3$. The simulation images are shown in Figure 3 and Figure 4.
\begin{figure}[t]
\centering
\includegraphics[width=6cm]{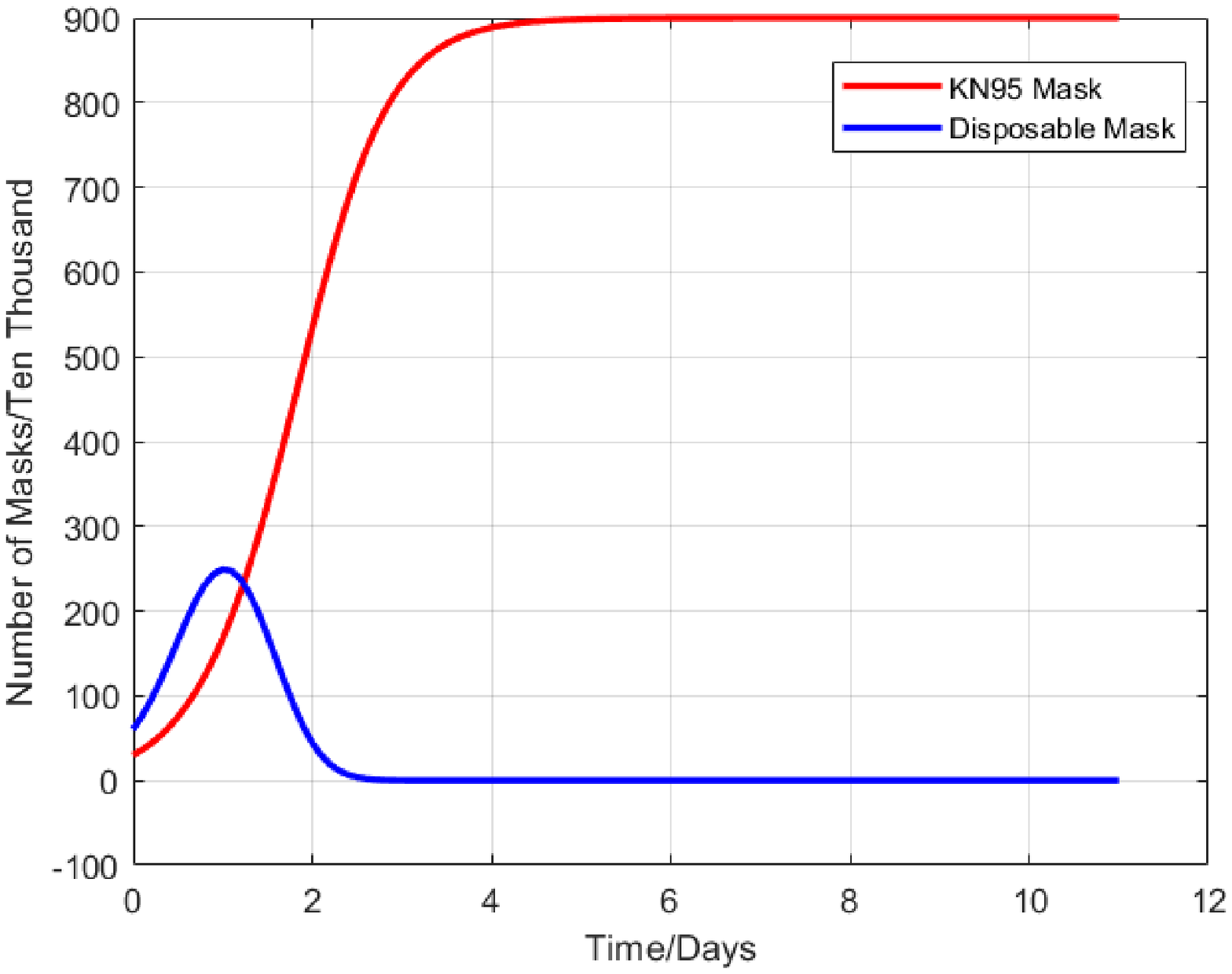}
\caption{Evolutionary Game of Two Types of Masks}
\label{fig}
\end{figure}

\begin{figure}[t]
\centering
\includegraphics[width=6cm]{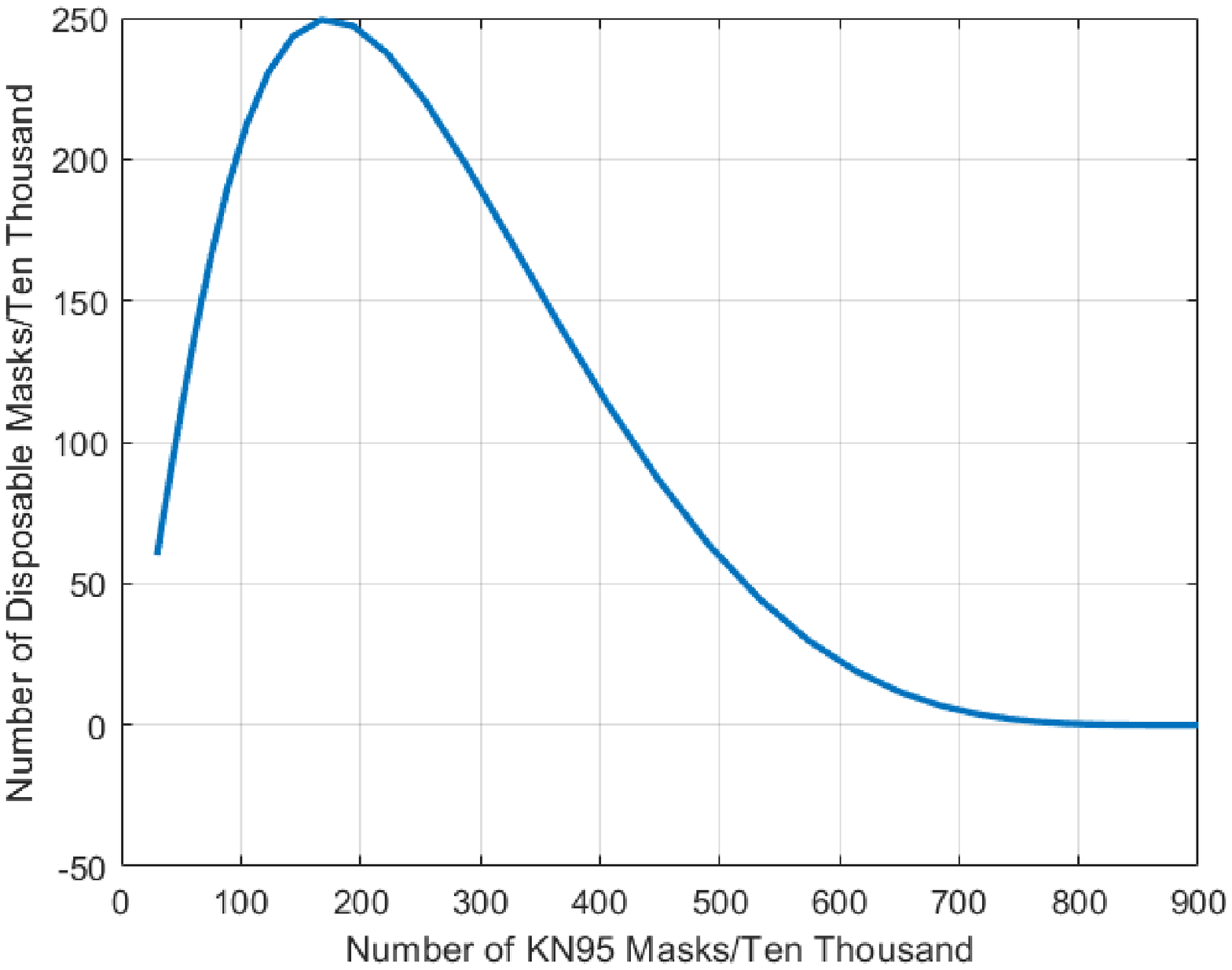}
\caption{Two Types of Masks Number Comparison}
\label{fig}
\end{figure}

Figure 3 shows the evolutionary game confrontation between KN95 masks and ordinary disposable medical masks in high-level industrial areas. One time unit before the outbreak of the epidemic, the output of disposable masks had an advantage in the market. At this time, the output of KN95 masks was steadily increasing. Between the first time unit and the 1.2th time unit, the number of KN95 masks was still rising, but the market share of disposable masks began to decline. After the 1.2th time unit, the number of KN95 masks has an overwhelming advantage, and the market share of disposable masks will be gradually replaced by KN95 masks.
From the above analysis, it can be seen that in high-level production areas, due to technological leadership, the production efficiency of KN95 masks has been significantly improved. The time for all regional personnel to use KN95 masks has been shortened from 9 unit time to 4 unit time, as soon as possible The local use of masks with higher levels of protection is more beneficial to the region’s fight against the epidemic. In this area, the production efficiency of KN95 masks is high, and when disposable medical masks reach 2 million in the area, the consumption is much greater than that of KN95 masks.
Manufacturers in this region can produce 9 million masks according to market demand when the epidemic breaks out to supply all personnel in the region.

\subsection{situation 3}
If the development of the regional mask industry is at a low level. The ability of regional masks to produce KN95 masks is lower than the medium level, that is, $r_{1}=0.5,r_{2}=3$. The simulation images are shown in Figure 5 and Figure 6.
\begin{figure}[t]
\centering
\includegraphics[width=6cm]{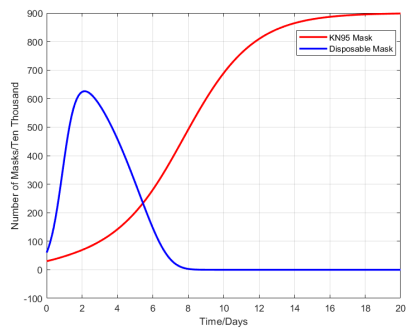}
\caption{Red and Blue Evolutionary Game Confrontation}
\label{fig}
\end{figure}

\begin{figure}[t]
\centering
\includegraphics[width=6cm]{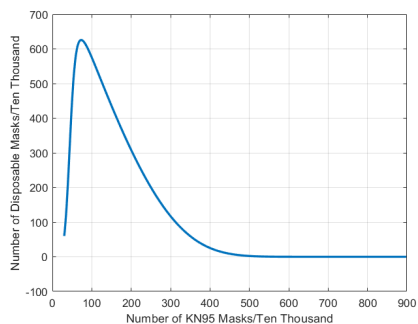}
\caption{Red and Blue Armor Number Comparison}
\label{fig}
\end{figure}
Figure 5 shows the evolutionary game confrontation between KN95 masks and ordinary disposable medical masks in low-level industrial areas. 2.1 time units before the outbreak, the output of disposable masks had an advantage in the market. At this time, the output of KN95 masks was steadily increasing. Between the 2.1th time unit and the 5.8th time unit, the number of KN95 masks was still rising, but the market share of disposable masks began to decline. After the 5.8th time unit, the number of KN95 masks has an overwhelming advantage, and the market share of disposable masks will be gradually replaced by KN95 masks.
The production efficiency of KN95 masks is significantly reduced in this area. The time for all regional personnel to use KN95 masks has been lengthened from 4 unit time to 18 unit time, which is extremely unfavorable in fighting the epidemic. At this time, a large amount of medical disposable masks should be invested in the early stage of the fight against the epidemic, and the output of disposable masks should be increased. When necessary, disposable masks should be purchased across regions. According to this situation, mask manufacturers produced a large number of disposable masks in the early stage of the fight against the epidemic and produced a small amount of KN95 masks. After the fight against the epidemic has experienced 6 time units, they can start producing KN95 masks. After that, they must always pay attention to the market demand for masks. Changes, KN95 masks are produced according to market demand changes.
In summary, the key factor to obtain the competitive advantage of KN95 masks and ordinary disposable medical masks is to objectively analyze market demand. According to market demand, KN95 masks and ordinary disposable medical masks are combined with the analysis of the evolutionary game competition model. KN95 masks can be made as the trend of regional personnel selection.

\section{CONCLUSION}
In this paper, based on the evolutionary game competition model of KN95 masks and ordinary disposable medical masks, the mathematical model is constructed using the principle of species competition mechanism in the evolutionary game algorithm, and the Runge-Kutta method is used to numerically solve the evolutionary differential equation. The evolutionary game competition between KN95 masks and ordinary disposable medical masks, and the corresponding relationship between the number and time of KN95 masks and ordinary disposable medical masks is comprehensively given. In this paper, the simulation advantage of the evolutionary game competition model of KN95 masks and ordinary disposable medical masks is that they can grasp the trend of mask production from a macro perspective, and provide important reference information for fighting the epidemic.

\end{document}